\documentclass[10pt,twocolumn,letterpaper]{article}

\usepackage{3dv}
\usepackage{times}
\usepackage{epsfig}
\usepackage{graphicx}
\usepackage{amsmath}
\usepackage{amssymb}
\usepackage{multirow}
\usepackage{threeparttable}

\usepackage[pagebackref=true,breaklinks=true,colorlinks,bookmarks=false]{hyperref}

\threedvfinalcopy 

\def\threedvPaperID{16} 

\ifthreedvfinal\fi
\begin{document}

\title{Deep Learning for Multi-view Stereo via Plane Sweep: A Survey}

\author{Qingtian Zhu$^1$, Chen Min$^{1,2}$, Zizhuang Wei$^1$, Yisong Chen$^1$, and Guoping Wang$^1$\\
$^1$Peking University\\
$^2$National Innovation Institute of Defense Technology\\
{\tt\small zqt@stu.pku.edu.cn}
}

\maketitle

\begin{abstract}
   3D reconstruction has lately attracted increasing attention due to its wide application in many areas, such as autonomous driving, robotics and virtual reality. As a dominant technique in artificial intelligence, deep learning has been successfully adopted to solve various computer vision problems. However, deep learning for 3D reconstruction is still at its infancy due to its unique challenges and varying pipelines. To stimulate future research, this paper presents a review of recent progress in deep learning methods for Multi-view Stereo (MVS), which is considered as a crucial task of image-based 3D reconstruction. It also presents comparative results on several publicly available datasets, with insightful observations and inspiring future research directions.
\end{abstract}

\section{Introduction}

With the rapid development of 3D acquisition techniques, depth sensors are becoming increasingly affordable and reliable, such as LiDARs. These sensors have been widely equipped for real-time tasks to obtain a rough estimation of surrounding environment, e.g., in simultaneous localization and mapping (SLAM). However, depth maps captured by depth sensors are usually sparse due to hardware and power limitations at edge devices, so that delicate details are abandoned in exchange for computational efficiency. Another pipeline is to reconstruct 3D models from a series of images. In this case, depth values are computed by matching 2D images and the whole reconstruction is done off-line. Provided that capturing images is more economical and available than acquiring depth maps via depth sensors, image-based 3D reconstruction is a better option for time-insensitive tasks. Besides, images actually contain information that depth sensors cannot capture, such as texture and lighting. These clues are crucial for reconstructing more delicate and detailed 3D models.

Multi-view Stereo (MVS) is a computationally expensive procedure of image-based 3D reconstruction. The most universal definition of the task is stated as follows. Given a series of images with their respective calibrated camera parameters, MVS aims to estimate a depth map for each image and then reconstruct a dense point cloud of the scene. Most previous attempts~\cite{yao2018mvsnet,yao2019recurrent} adopt this definition. Deep learning has shown its effectiveness in many computer vision tasks. For binocular stereo, \cite{okutomi1993multiple} discretizes the depth space and turns the task of stereo into a classification problem. Plane sweep algorithm~\cite{collins1996space} extends this pattern to multi-image matching, whose pattern is suitable for deep CNNs to handle.

\begin{figure*}
    \centering
    \includegraphics[width=\linewidth]{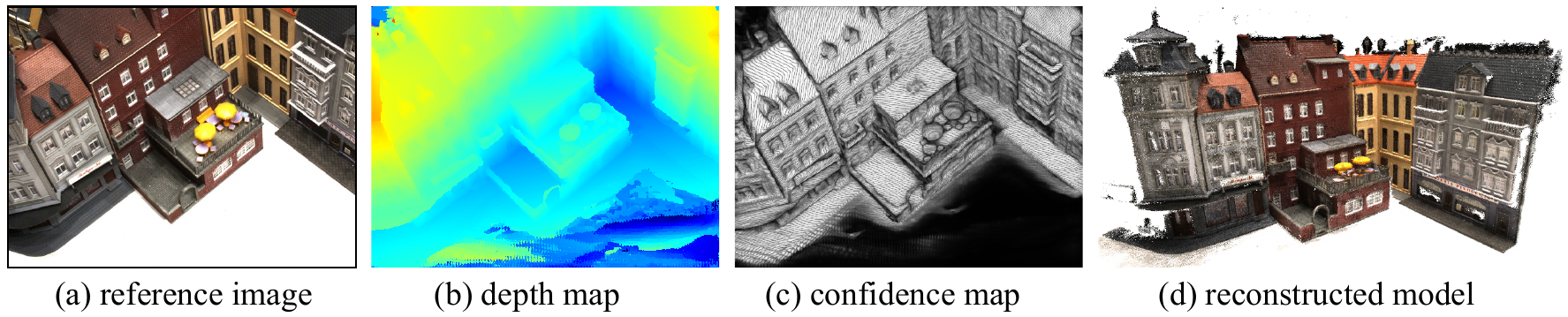}
    \caption{Input, intermediate outputs and final output from a typical MVS pipeline. Reference image (a), together with its $N-1$ neighboring images, are the inputs of network and its depth map (b) and confidence map (c) are produced accordingly. Depth maps of all images are filtered and fused into the reconstructed dense point cloud (d).}
    \label{fig:overall}
\end{figure*}

To restate, this paper surveys learning-based MVS methods that build matching cost volumes via plane sweep algorithm and yield per-view depth maps as intermediate representation to reconstruct dense 3D point clouds. Each image within a scan takes turns to be the reference image to estimate depth maps and its $N-1$ neighboring images as source images. In total $N$ images are sent into the network as inputs to produce one depth map and corresponding confidence map. The dense point cloud is then obtained by filtering and fusing depth maps for all images, The pipeline is also shown in Fig.~\ref{fig:overall}.

\section{Background}
MVS plays as a key component of image-based 3D reconstruction pipeline. In this section, background knowledge is presented. Sec.~\ref{sec:sfm} introduces preproposed Structure from Motion (SfM) as the source of camera calibration. Sec.~\ref{sec:ps} explains how to build cost volumes in learning-based MVS methods, which is a key step to enable CNNs to predict depth. Post-processes after obtaining depth maps, including depth filter and fusion, are presented in Sec.~\ref{sec:dff}. Sec.~\ref{sec:datasets} lists several well-known open datasets for MVS and Sec.~\ref{sec:em} lists metrics used for evaluation. Sec.~\ref{sec:loss} covers loss functions used for learning-based MVS. For clarification, this survey only cover learning-based MVS methods taking advantage of plane sweep algorithm.

\subsection{Structure from Motion}\label{sec:sfm}
MVS requires calibrated camera parameter to obtain image-wise adjacency, which is usually achieved by Structure from Motion algorithms. SfM is usually categorized into incremental and global ones. Generally speaking, incremental pipelines solve the optimization problem locally and merge new cameras into known tracks. Thus incremental methods are slower but more robust and accurate. Global SfM is more scalable and can often converge to a pretty good solution but is more susceptible to outliers.

Specifically for MVS, camera calibration means for each image, a camera extrinsic matrix $\mathbf{T}$, a camera intrinsic matrix $\mathbf{K}$, a depth range $[d_{min},d_{max}]$ are acquired by SfM. For most MVS methods, COLMAP~\cite{schonberger2016structure} provides a good enough estimation of cameras. 

\subsection{Plane Sweep}\label{sec:ps}
\begin{figure}
    \centering
    \includegraphics[width=\linewidth]{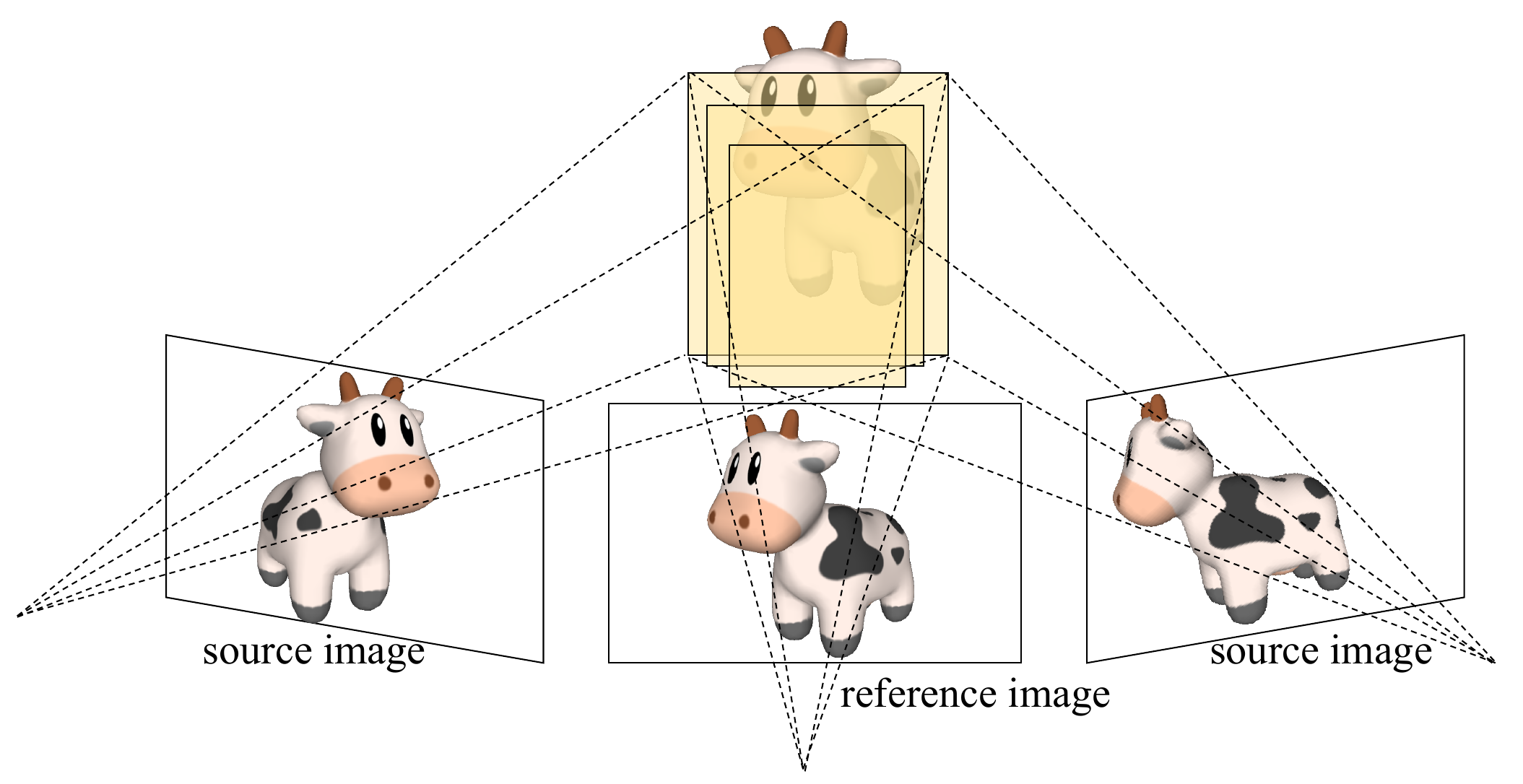}
    \caption{Illustration of plane sweep algorithm. To estimate the depth map for reference camera, neighboring source images are projected by homography to fronto-parallel planes of the frustum of the reference camera.}
    \label{fig:sweep}
\end{figure}
The main principle of plane sweep stereo~\cite{collins1996space} is that for each depth, source images are projected to fronto-parallel planes of the reference camera frustum and those depth hypotheses with high similarity of projected images are more reliable. Most learning-based MVS methods rely on plane sweep algorithm to generate cost volumes. This practice is deeply inspired by binocular stereo. In learning-based binocular stereo methods, instead of regressing depth values directly, disparity values, which describe the pixel-level distance between the two views, are estimated. With knowledge of epipolar geometry, depth values can be computed from estimated disparity values. Besides, since the unit of disparity values are pixels, this task becomes a classification task where each class represents a discretized disparity. This common practice has two underlying advantages. First, the estimation of depth is now scale-irrelevant since the unit is pixels, not meters or other measurement of actual distance. Second, CNNs are considered as better at classification than regression, so this helps to yield more reliable results. This discretization relies on plane sweep algorithm.

The core of plane sweep algorithm is to verify depth hypotheses. After projecting pixels into space by a hypothetical depth, plane sweep algorithm says that if a hypothetical point in space is captured by different cameras with a similar photometry, this point is likely to be a real point, which is to say, the hypothesis of depth ($z$ value) is valid. In this case, we can divide the depth interval into discretized values and make hypotheses with these values. The final depth is estimated by choosing the most valid depth among all hypotheses. When it comes to implementation, there are two problems remaining. One is to match pixels across different images or to build a homography between views; the other is to measure the similarity of photometry. Note that considering individual pixel RGB color is not robust enough for matching, photometry is usually replaced by a feature map extracted from the original image.

For a pair of calibrated binocular images, since the two main optical axes are always parallel, we only need to shift one view to another by disparity hypotheses. Things get a little more complicated for MVS as cameras are distributed over the space without epipolar constraint. At depth hypothesis $d$, we first project all pixels of a source image into space with $d$ and then back-warp these points through the reference camera. Thus the homography between the $i$-th source image and the reference image is
\begin{equation}\label{equ:homo}
    H_i(d) = d\mathbf{K}_0\mathbf{T}_0\mathbf{T}^{-1}_i\mathbf{K}^{-1}_i,
\end{equation}
where $\mathbf{K}_0$ and $\mathbf{T}_0$ are camera intrinsics and extrinsics of reference image.

Measurement of photometric similarity varies from one method to another. For binocular stereo, \cite{dosovitskiy2015flownet,mayer2016large} introduce a correlation layer to compute inner product of feature vectors; GC-Net~\cite{kendall2017end} concatenates feature vectors together. For MVS where the number of cameras is larger than two, there are two main options. MVSNet~\cite{yao2018mvsnet} applies variance for all feature vectors; DPSNet~\cite{im2018dpsnet} concatenates features pairwise and a final cost volume is obtained by averaging all $N-1$ volumes.

It is worth noting that the division of depth space is a crucial problem to yield good results, which will be covered later in Sec.~\ref{sec:loss}.

To restate with an example, provided that the image resolution is $H\times W$ and the number of total depth hypotheses is $D$, assuming the dimension of per-pixel feature vectors is $F$, MVS methods build a cost volume of $H\times W\times D\times F$ from image features and this cost volume is then regularized by a neural network to obtain a depth map.

\subsection{Depth Filtering \& Fusion}\label{sec:dff}
Assuming that all depth maps have been obtained by MVS methods, the next step is to filter and fuse depth maps into a dense point cloud.
Since image-based 3D reconstruction is scale-irrelevant, the estimated depth values are actually the $z$ values for pixels in the local camera coordinate system. Thus the fusion of depth maps are rather straightforward that all we need to do is to project all pixels into 3D space through cameras. The transformation between image coordinate and world coordinate is
\begin{equation}
    \mathbf{P}_{w} = d\mathbf{T}^{-1}\mathbf{K}^{-1}\mathbf{P}_{x},
\end{equation}
where $\mathbf{P}_{x}$ and $\mathbf{P}_{w}$ denote pixel coordinates in image coordinate and world coordinate respectively.

However, not all pixels are suitable to be preserved in the final point cloud, e.g., those with low confidence and those at infinity, such as sky. To overcome this problem, depth maps are filtered before fused. Since learning-based MVS methods adopt a classification fashion, each depth map is yielded together with a confidence map correspondingly. So naturally, a threshold can be set to filter depth values with low confidence. Besides, depth values can be across-checked among neighboring views. This strategy of filtering is based on reprojection error, which is commonly used in Bundle Adjustment of SfM.

\begin{figure}
    \centering
    \includegraphics[width=0.9\linewidth]{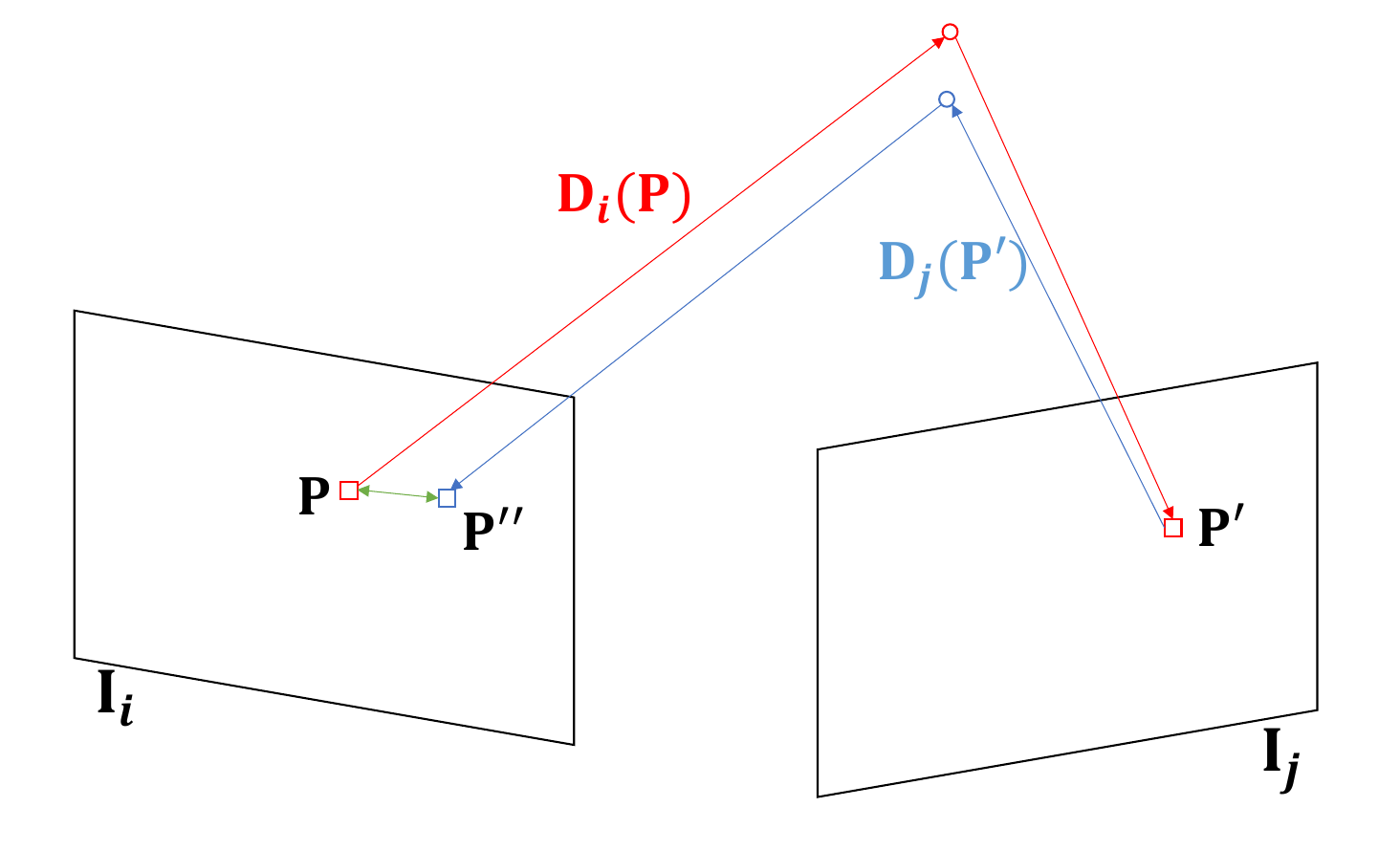}
    \caption{Illustration of checking geometric consistency among neighboring views by measuring reprojection error.}
    \label{fig:geo}
\end{figure}

Taking the strategy in \cite{yi2020pyramid} as an example, which is also shown in Fig.~\ref{fig:geo}, by mapping a pixel $\mathbf{P}$ in image $\mathbf{I}_i$ to its neighboring view $\mathbf{I}_j$ through estimated depth $D_i(\mathbf{P})$, we obtain a new pixel $\mathbf{P}'$. As $\mathbf{I}_j$ also has its depth map, we can get $D_j(\mathbf{P}')$ accordingly. In turn, $\mathbf{P}'$ can be projected to $\mathbf{I}_i$ at $\mathbf{P}''$ with depth $D_j(\mathbf{P}')$. Depth estimation of $\mathbf{P}''$ in $\mathbf{I}_i$ is denoted as $D_i(\mathbf{P}'')$.
The constraints for depth filtering are 
\begin{equation}
    \| \mathbf{P}-\mathbf{P}''\|_2\leq \tau_1,
\end{equation}
\begin{equation}
   \frac{\|D_i(\mathbf{P}'')-D_i(\mathbf{P})\|_1}{D_i(\mathbf{P})}\leq \tau_2,
\end{equation}
where $\tau_1$ and $\tau_2$ are threshold values. As for \cite{yi2020pyramid}, pixels satisfying these constraints under at least 3 neighboring views are considered as valid enough to remain.
It is worth noting that depth filtering and fusion methods are often uncovered in papers though they might be of great importance to obtain good results.

\subsection{Datasets}\label{sec:datasets}

Tab.~\ref{tab:datasets} is a brief summary of released MVS datasets. Note that for MVS training, depth is required while evaluation is based on point clouds. Surface reconstruction is required to render depth maps from point clouds and depth fusion is required to evaluate reconstruction quality. Besides, if a dataset does not consist of ground truth camera calibration or uses an open-source software to obtain ground truth calibration, then it might not be suitable for training since plane sweep is sensitive to noises in camera calibration.
\begin{table*}[t]
    \caption{An overview of public datasets for MVS.}
    \centering
    \resizebox{0.85\textwidth}{!}{ 
    \begin{threeparttable}
    {
    \begin{tabular}{|l|c|c|c|c|c|c|}
    \hline
    \multirow{2}{*}{\textbf{Dataset}} & \multicolumn{3}{c|}{\textbf{Provided Ground Truth}\tnote{1}}  & \multirow{2}{*}{\textbf{Synthetic}} & \textbf{Online} & \textbf{Evaluation} \\ 
    \cline{2-4}
    & Camera & Depth & Point Cloud  &  & \textbf{Benchmark} & \textbf{Target}\\
    \hline
    DTU~\cite{aanaes2016large} & \checkmark &  & \checkmark &  &  & Point Cloud \\
    \hline
    Tanks and Temples~\cite{knapitsch2017tanks} &  &  & \checkmark &  & \checkmark & Point Cloud \\
    \hline
    ETH3D~\cite{schops2017multi} & \checkmark  &   & \checkmark &  & \checkmark & Point Cloud \\
    \hline
    BlendedMVS~\cite{yao2020blendedmvs} & \checkmark & \checkmark & & \checkmark &  & Depth Map\\ 
    \hline
    \end{tabular}
    \begin{tablenotes}
     \item[1] For datasets with online benchmark, ground truth of test set (excluding camera parameters) is not released.
   \end{tablenotes}
    }
    \end{threeparttable}
    }
\label{tab:datasets}
\end{table*}

\paragraph{DTU} 
DTU dataset~\cite{aanaes2016large} is an indoor MVS dataset collected under well-controlled laboratory conditions with accurate camera trajectory. It contains 128 scans with 49 views under 7 different lighting conditions and is split into 79 training scans, 18 validation scans and 22 evaluation scans. By setting each image as reference, there are 27097 training samples in total. DTU dataset officially provides ground truth point clouds, rather than depth maps, which means surface reconstruction is required to generate mesh models and render depth maps. Normally, screened Poisson surface reconstruction algorithm~\cite{kazhdan2013screened} is adopted.

\paragraph{Tanks and Temples} 
Tanks and Temples~\cite{knapitsch2017tanks} is a large-scale online benchmark captured in more complex real indoor and outdoor scenarios. It contains an intermediate set and an advanced set. Different scenes have different scales, surface reflection and exposure conditions. Evaluation of Tanks and Temples is done online by uploading reconstructed points to its official website. Note that Tanks and Temples does not provide ground truth camera parameters. A training set with ground truth point clouds is available, which is usually used for local off-line validation.

\paragraph{ETH3D}
ETH3D~\cite{schops2017multi} is a comprehensive benchmark for both SLAM and stereo tasks. Considering MVS, it contains 25 high-resolution scenes and 10 low-resolution scenes. ETH3D is widely acknowledged as the most difficult MVS task since it contains many low-textured regions such as white walls and reflective floor. Traditional MVS methods based on broadcasting valid depth values perform better in this case.

\paragraph{BlendedMVS} 
BlendedMVS dataset~\cite{yao2020blendedmvs} is a recently published large-scale synthetic dataset for MVS training that contains a variety of scenes, such as cities, sculptures and shoes. The dataset consists of over 17k high-resolution images rendered with reconstructed models and is split into 106 training scenes and 7 validation scenes. Since BlendedMVS is obtained through virtual cameras, its provided camera calibration is reliable enough for MVS training.

\subsection{Evaluation Metrics}\label{sec:em}
As is mentioned in Tab.~\ref{tab:datasets}, most datasets provide point clouds as ground truth instead of depth maps and evaluation metrics are usually based on quality of reconstructed dense point clouds. Since point clouds are actually unordered points with permutation invariance, before comparison, reconstructed point clouds should be aligned to ground truth point clouds through per-view camera parameters.

\paragraph{Absolute Error}
Though benchmarks adopt point-cloud-based metrics for ranking, depth-based metrics can still be used for validation during network training. Absolution error is commonly used to measure the quality of depth maps. A common practice is to use multiple thresholds to show a more overall performance of networks, e.g., 2-px absolute error, 4-px absolute error, 6-px absolute error and 8-px absolute error.

\paragraph{Precision/Accuracy}
Precision/Accuracy is a measurement of what percentage of predicted points can get matched in the ground truth point cloud. Considering a point $\mathbf{P}_p$ in the predicted point cloud, it is considered to have a good match in the ground truth point cloud $\{\mathbf{P}_g\}$ if
\begin{equation}
    \|\mathbf{P}_p - \mathop{\arg\min}\limits_{\mathbf{P}\in \{\mathbf{P}_g\}} \|\mathbf{P}-\mathbf{P}_p\|_2\|_2 \leq \lambda,
\end{equation}
where $\lambda$ is a scene-dependent parameter assigned by datasets. $\lambda$ is set to a larger value for larger scenes. The definition of distance is the same as Chamfer distance. Precision/Accuracy is the number of points in the predicted point cloud satisfying the requirement over the total number of points in the predicted point cloud. Note that in some datasets, precision/accuracy is not measured by proportion (percentage) but by mean or median absolute distance instead.

\paragraph{Recall/Completeness}
Recall/Completeness measures what percentage of ground truth points can get matched in the predicted point cloud. The computation is simply swapping the ground truth and the prediction. For a point $\mathbf{P}_g$ in the ground truth point cloud, it is considered to have a good match in the predicted point cloud $\{\mathbf{P}_p\}$ if
\begin{equation}
    \|\mathbf{P}_g - \mathop{\arg\min}\limits_{\mathbf{P}\in \{\mathbf{P}_p\}} \|\mathbf{P}-\mathbf{P}_g\|_2\|_2 \leq \lambda,
\end{equation}
and recall/completeness is the number of points in the ground truth point cloud satisfying the requirement over the total number of points in the ground truth point cloud. Similar to precision/accuracy, recall/completeness is sometimes measured by mean or median absolute distance.

\paragraph{F-Score}
The two aforementioned metrics measure the accuracy and completeness of predicted point clouds. However, each of these metrics alone cannot present the overall performance since different MVS methods use different a prior assumptions. A stronger assumption usually leads to higher accuracy but lower completeness. 
Both measures are needed for a fair comparison. If only precision/accuracy is reported, it would favor MVS algorithms that only include estimated points of high certainty. On the other hand, if only recall/completeness is reported it would favor MVS algorithms that include everything, regardless of point quality.
Therefore, an integrated metric is introduced. F-score is the harmonic mean of precision and recall. Harmonic mean is sensitive to extremely small values and tends to get more affected by smaller values, which is to say, F-score does not encourage imbalanced results. However, in most cases, F-score still suffers from unfairness due to limitations of ground truth. Since the representation of point clouds is unstructured and overall sparse, this problem remains unsolved.

\subsection{Loss Function}\label{sec:loss}
Loss functions for learning-based MVS can be categorized into regression-like and classification-like ones. To briefly recap, for a cost volume of $H\times W\times D\times F$, a probability volume $H\times W\times D$ is generated after cost volume regularization. Different loss functions correspond to different ways of determining the final prediction.

If the final ground truth prediction is determined by an argmax operation. It has already been turned into a pure classification task, cross entropy loss is naturally suitable to be the loss function, where the ground truth depth maps are also discretized in the same way of plane sweep and one-hot encoded. The cross entropy loss function is stated as
\begin{equation}
    L = \sum_d^D -G(d)\log[P(d)],
\end{equation}
where $G(d)$ is the ground truth one-hot distribution w.r.t. depth and $P(d)$ is the predicted distribution. An important advantage of classification-like loss functions are actually insensitive to depth division, which means the division can be arbitrary and not necessarily to be uniform. \cite{yao2019recurrent,yan2020dense} use cross entropy as their loss functions.

Some methods adopt a regression-like pattern to determine the prediction that the mathematical expectation of depth is calculated instead. In this case, a L1 loss is adopted as the loss function. This practice helps to predict smoother depth maps. MVSNet~\cite{yao2018mvsnet}, along with later coarse-to-fine methods, adopts this pattern. The loss is stated as
\begin{equation}
    L = \|d_0 -Ex[P(d)]\|_1
\end{equation}
where $d_0$ denotes the ground truth depth map and $Ex[\cdot]$ denotes expectation of a distribution. However, mathematical expectation is valid if and only if the division of space is uniform. To boost the scalability, R-MVSNet~\cite{yao2019recurrent} adopts an inverse depth sampling strategy where the level of depth planes and actual depth value are in inverse proportion. In this case, plane sweep is finer at distant areas but the regression-like loss function is no longer valid.

Empirical results show that classification-like loss functions help to predict accurate depth values since all candidates lower than maximum probability are compressed. However, it usually leads to incontinuity of depth values.
While regression-like ones consider mathematical expectation is a differentiable way to do argmax, which helps to predict smooth depth maps but loses sharpness on edges.

\section{Method}\label{sec:methods}
This section introduces learning-based MVS networks that yield depth maps for further post-processing. A typical MVS network mainly contains three parts, namely a feature extraction network (Sec.~\ref{sec:fe}), a cost volume constructor (Sec.~\ref{sec:cvc}) and a cost volume regularization network (Sec.~\ref{sec:cvr}). Tab.~\ref{tab:methods} is an overview of typical learning-based MVS methods.

\begin{table*}[t]
    \caption{An overview of typical learning-based MVS methods via plane sweep algorithm.}
    \centering
    \resizebox{0.85\textwidth}{!}{ 
    \begin{threeparttable}
    {
    \begin{tabular}{|l|c|c|c|c|c|c|}
    \hline
    \multirow{2}{*}{\textbf{Method}} & \multicolumn{3}{c|}{\textbf{Regularization Scheme}\tnote{1}}  & \multirow{2}{*}{\textbf{Visibility}} & \multicolumn{2}{c|}{\textbf{Loss Function}}  \\ 
    \cline{2-4} \cline{6-7}
    & 3D CNN & RNN & Coarse to Fine  &  & Classification & Regression\\
    \hline
    MVSNet~\cite{yao2018mvsnet} & \checkmark &  &  &  &  & \checkmark \\
    \hline
    R-MVSNet~\cite{yao2019recurrent} &  & \checkmark &  &  & \checkmark &  \\
    \hline
    CasMVSNet~\cite{gu2020cascade} & & &\checkmark & &  & \checkmark\\
    \hline
    CVP-MVSNet~\cite{yang2020cost} & & &\checkmark & &  & \checkmark\\
    \hline
    UCS-Net~\cite{cheng2020deep} & & &\checkmark & &  & \checkmark\\
    \hline
    Vis-MVSNet~\cite{zhang2020visibility} & & &\checkmark & \checkmark & \checkmark & \\
    \hline
    PVA-MVSNet~\cite{yi2020pyramid} & \checkmark &  &  & \checkmark &  & \checkmark \\
    \hline
    $D^2$HC-RMVSNet~\cite{yan2020dense} &  & \checkmark &  &  & \checkmark &  \\
    \hline
    AA-RMVSNet~\cite{wei2021aa} &  & \checkmark &  &  \checkmark & \checkmark &  \\
    \hline
    \end{tabular}
    \begin{tablenotes}
     \item[1] Strictly speaking, all coarse-to-fine methods are all based on 3D CNNs. Therefore, the definition of using 3D CNN excludes those with a coarse-to-fine pattern.
   \end{tablenotes}
    }
    \end{threeparttable}
    }
\label{tab:methods}
\end{table*}

\subsection{Feature Extraction}\label{sec:fe}
Feature extraction for MVS still remains unstudied and most methods apply a common CNN backbone methods to extract features, e.g., ResNet~\cite{he2016deep} and U-Net~\cite{ronneberger2015u}. The main novelty of learning-basedlies at feature extraction, e.g., $D^2$HC-RMVSNet~\cite{yan2020dense} applies multi-scale features with dilation to aggregate features.

We can compare feature extraction of different computer vision tasks. For image classification where each image is assigned with one label, global features are more important since an overall perception of the whole image is required. For object detection, locality is more significant than global context. As for stereo matching, which is rather similar to MVS, the best matching should be semi-global~\cite{hirschmuller2007stereo}. For high-frequency regions whose texture information is rich, we expect a more local receptive field; while those weak-textured areas should be matched in a wider range.

\subsection{Cost Volume Construction}\label{sec:cvc}
Plane sweep is done to construct the cost volume, whose details have been presented in Sec.~\ref{sec:ps}. Since cost volume construction can be pairwise, there occurs another procedure to aggregate all $N-1$ cost volumes into one. DPSNet~\cite{im2018dpsnet} simply aggregates all cost volumes by addition and the underlying principle is that all views are considered equally. Practically speaking, occlusion is common in a MVS system and it usually causes invalid matching. As a result, an increasing number of input views will lead to even worse prediction. In this way, views that are closer to the reference view should be given higher priority since it is less likely to suffer from occlusion.

To alleviate this problem, PVA-MVSNet~\cite{yi2020pyramid} applies gated convolution~\cite{yu2019free} to adaptively aggregate cost volumes. View aggregation tends to give occluded areas smaller weights and the reweighting map is yielded according to the volume itself. This practice actually follows the fashion of self-attention. Vis-MVSNet~\cite{zhang2020visibility} explicitly introduces a measure for cost volumes treated as visibility by examining the uncertainty or confidence of probability distribution.

\subsection{Cost Volume Regularization}\label{sec:cvr}
\begin{figure*}
    \centering
    \includegraphics[width=\linewidth]{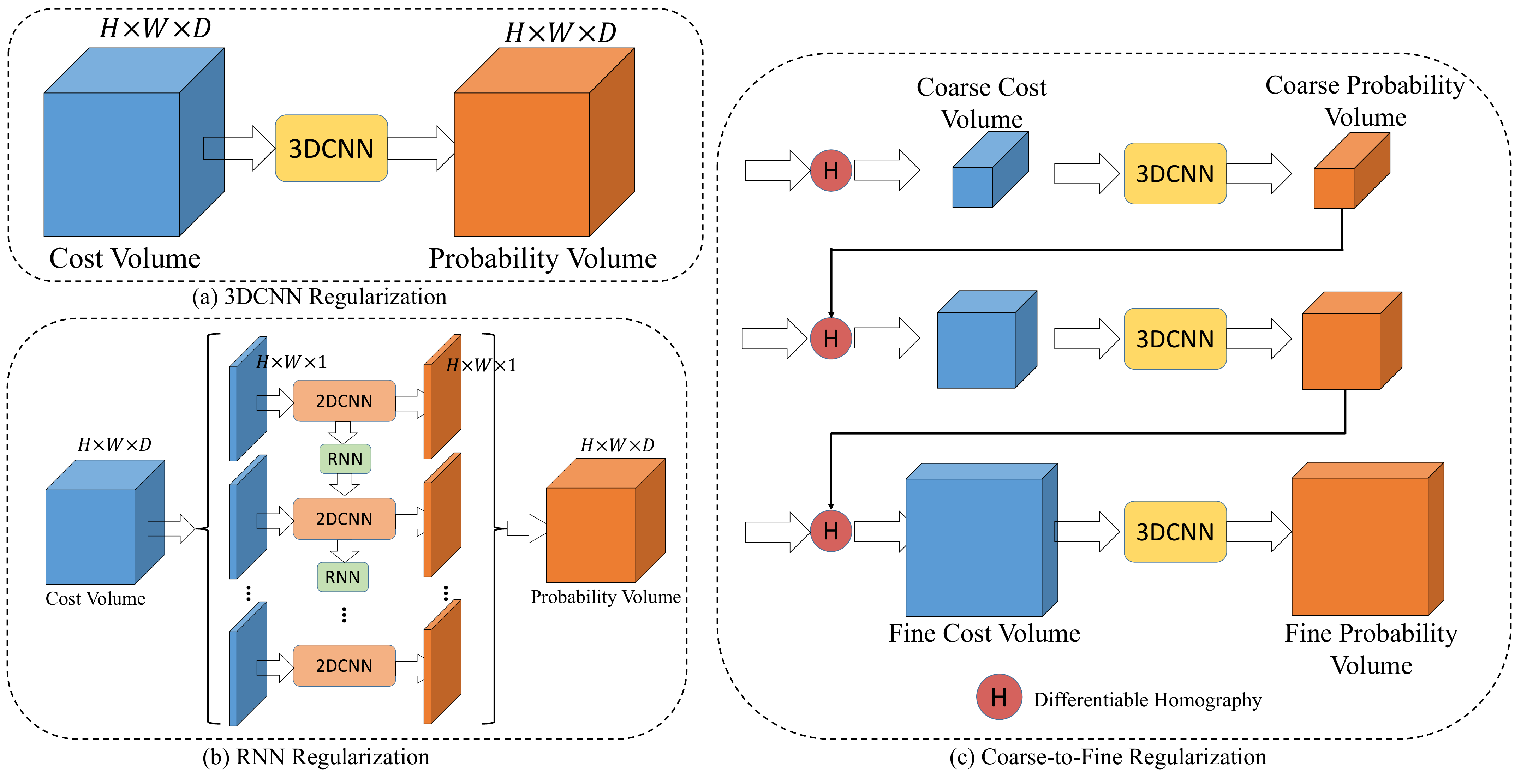}
    \caption{Illustration of typical cost regularization schemes. (a) 3D CNN regularization simply applies a 3D CNN to aggregate spatial context of all dimensions; (b) RNN regularization models each depth hypothesis similarly to a time node and adopts a 2D CNN with shared weights to aggregate context of $H$ and $W$; (c) in coarse-to-fine regularization, finer cost volumes are built according to coarser prediction and 3D CNNs are used for different stages.}
    \label{fig:regularization}
\end{figure*}

The main differences between different MVS networks lie in the way of doing cost volume regularization, which will be categorized and introduced in the following sections. Fig.~\ref{fig:regularization} illustrates the three regularization schemes covered in this paper.

\subsubsection{3D CNN}
3D CNNs are a straightforward choice for cost volume regularization. Literally, 3D CNNs consist of 3D convolution operations, whose convolution kernels are 3 dimensional and move among all dimensions of cost volumes. MVSNet~\cite{yao2018mvsnet} adopts a 3D U-Net to regularize cost volumes. Similar to 2D U-Net~\cite{ronneberger2015u}, 3D U-Net contains an encoder, which does downsampling 3D convolutions, and a decoder, that recovers the original feature resolution gradually. MVSNet~\cite{yao2018mvsnet}, which is the first MVS method that leverages deep learning, uses a 3D U-Net for cost volume regularization.

The purpose of cost volume regularization is to aggregate features and predict relatively valid depth values according to aggregated features. In this way, 3D CNNs are a universal method for its capability of aggregating local and global features in all dimensions. However, CNNs are operated on regular grids and a underlying assumption is that the division of space is uniform. For completed cases, a uniform division is not good enough to predict reliable depth values. Besides, 3D CNNs are computationally expensive and consumes massive memory, which limits the value of $D$. A finer (large $D$) and suitable (not uniform) division is usually crucial to obtain high quality depth maps.  

\subsubsection{RNN}
A major disadvantage of using 3D CNNs for cost volume regularization is massive memory consumption and in order to reduce the amount of demanded memory, some attempts~\cite{yao2019recurrent,yan2020dense} replace 3D CNNs by sequential 2D CNNs along the dimension of $D$. In this way, there will always be one cost slice being processed in the GPU memory and a RNN is used to thread all depth hypotheses, passing context information on dimension $D$.

A great advantage of using recurrent regularization is that it improves the scalability of MVS methods since the division of space can be finer so farther objects can be reconstructed. But correspondingly, this scheme trades running efficiency for space since RNN's parallelizability is poorer than CNNs.

\subsubsection{Coarse to Fine}
Making prediction in a coarse-to-fine pattern is another solution of reducing massive memory consumption. Literally, the network predicts a coarse depth map and then finer results are yielded based on former ones. Coarse predictions are usually based on downsampled images of lower resolution and fine predictions are on higher resolution images. This practice adopts encoder-decoder architectures where low-res feature maps contain more low frequency components and high frequency is more found in high-res feature maps. The ways of making use of coarse prediction are different. Cas-MVSNet~\cite{gu2020cascade} regenerates cost volumes by warping feature maps with a smaller depth range around previous coarse prediction. Cascade rewarping at stage $k+1$ is based on Eq.~\ref{equ:homo} by setting $d = d^{(k)}+\Delta^{(k+1)} $, whose homography is
\begin{equation}
    H_i^{(k+1)}(d^{(k)}+\Delta^{(k+1)}) = (d^{(k)}+\Delta^{(k+1)})\mathbf{K}_0\mathbf{T}_0\mathbf{T}^{-1}_i\mathbf{K}^{-1}_i,
\end{equation}
where $k$ denotes the number of stages, $d^{(k)}$ is the estimated depth value at stage $k$ and $\Delta^{(k+1)}$ is the residual depth to be determined in the current stage.

UCS-Net~\cite{cheng2020deep} obtains uncertainty from coarse prediction to aid finer prediction. Note that both RNN and coarse-to-fine regularization methods allow finer division of depth but they focus on different cases. RNN regularization allows larger $D$ so there can be more hypothetical depth planes; coarse-to-fine regularization enables an adaptive subdivision of depth interval for finer prediction, leading to ability of constructing delicate details.

\section{Discussions}

\subsection{Results \& Analysis}
Tab.~\ref{tab:dtu} and Tab.~\ref{tab:tnt} are quantitative results of typical learning-based MVS methods. Furu~\cite{furukawa2009accurate}, Gipuma~\cite{galliani2015massively} and COLMAP~\cite{schonberger2016pixelwise} are non-learning methods shown for comparison.

As for DTU dataset, it is apparent that learning-based MVS methods outperform traditional methods especially in terms of completeness, while Gipuma~\cite{galliani2015massively}, a non-learning method, achieves best accuracy. A straightforward insight is that deep learning can adopt a data-driven statistical pattern to predict complete depth values, while traditional methods depend on stricter constraints so that these regions are omitted due to poor geometric consistency. Among learning-based methods, coarse-to-fine ones perform better than others. Considering DTU is an ideal indoor dataset with delicate details, coarse-to-fine methods are able to distinguish subtle depth differences.

While for Tanks and Temples, whose scale is larger, methods using RNN regularization scheme demonstrate their dominance, e.g., $D^2$HC-RMVSNet~\cite{yan2020dense} and AA-RMVSNet~\cite{wei2021aa}. For real-world scenes, whose lighting and scale can be varying, it is more important to reconstruct relatively complete point clouds rather than accurate ones. Therefore, RNN regularization, which allows a larger number of $D$ and inverse depth sampling, is better at predicting complete scenes. 

This performance gap between DTU and Tanks and Temples can be explained from the perspective of regularization. 3D CNN (including coarse-to-fine regularization pattern) is a rather stronger regularization since it aggregates spatial context information in all dimensions, while in RNNs, the constraints in the dimension of $D$ are relaxed. The relaxation leads to better generalizability and robustness but takes more time for prediction.

\begin{table*}[t]
    \caption{Quantitative results on DTU evaluation set~\cite{aanaes2016large} (lower is better). Acc. stand for accuracy and Comp. stand for completeness. Note that DTU dataset measures absolute Chamfer distance between point clouds instead of percentage for evaluation. Furu~\cite{furukawa2009accurate}, Gipuma~\cite{galliani2015massively} and COLMAP~\cite{schonberger2016pixelwise} are non-learning methods listed for comparison.}
    \centering
    \begin{tabular}{|l|c|c|c|}
    \hline
    \textbf{Method} & \textbf{Acc.(mm)} & \textbf{Comp.(mm)} & \textbf{Overall(mm)} \\
    \hline
    Furu~\cite{furukawa2009accurate} & 0.613 & 0.941 & 0.777 \\ 
    \hline
    Gipuma~\cite{galliani2015massively} & 0.283 & 0.873 & 0.578 \\
    \hline
    COLMAP~\cite{schonberger2016pixelwise} & 0.400 & 0.664 & 0.532 \\ 
    \hline
    MVSNet~\cite{yao2018mvsnet} & 0.396 & 0.527 & 0.462 \\
    \hline
    R-MVSNet~\cite{yao2019recurrent} & 0.385 & 0.459 & 0.422 \\
    \hline
    $D^2$HC-RMVSNet~\cite{yan2020dense} & 0.395 & 0.378 & 0.386 \\
    \hline
    Vis-MVSNet~\cite{zhang2020visibility} & 0.369  & 0.361 & 0.365 \\
    \hline
    CasMVSNet~\cite{gu2020cascade} & 0.325 &  0.385 & 0.355 \\ 
    \hline
    CVP-MVSNet~\cite{yang2020cost} & 0.296 &  0.406 & 0.351 \\ 
    \hline
    UCS-Net~\cite{cheng2020deep} & 0.338 & 0.349 & 0.344 \\
    \hline
    AA-RMVSNet~\cite{wei2021aa} & 0.376 & 0.339 & 0.357 \\
    \hline
    \end{tabular}
    \label{tab:dtu}
\end{table*}
\begin{table*}[t]
    \caption{Quantitative results of typical learning-based MVS methods on the Tanks and Temples benchmark~\cite{knapitsch2017tanks}. The evaluation metric is F-score (higher is better). L.H. stands for Lighthouse and P.G. stands for Playground. COLMAP~\cite{schonberger2016pixelwise} is a non-learning baseline for comparison.}
    \centering
    \resizebox{0.85\textwidth}{!}{ 
    \begin{tabular}{|l|c|c|c|c|c|c|c|c|c|}
    \hline
    \textbf{Method} & \textbf{Mean} & Family & Francis & Horse & L.H. & M60 & Panther & P.G. & Train \\
    \hline
    COLMAP~\cite{schonberger2016pixelwise}& 42.14	&	50.41&	22.25&	25.63&	56.43&	44.83&	46.97&	48.53&	42.04\\
    \hline
    MVSNet~\cite{yao2018mvsnet}&	43.48&	55.99&	28.55&	25.07&	50.79&	53.96&	50.86&	47.90&	34.69\\
    \hline
    R-MVSNet~\cite{yao2019recurrent}&	50.55&	73.01&	54.46&	43.42&	43.88	&46.80&	46.69&	50.87&	45.25\\
    \hline
    PVA-MVSNet~\cite{yi2020pyramid}&  54.46	&	69.36 & 46.80 &	46.01 &	55.74 &	57.23 &	54.75 &	56.70 &	49.06 \\
    \hline
    CVP-MVSNet~\cite{yang2020cost} &	54.03&	76.50&	47.74&	36.34&	55.12&	57.28&	54.28&	57.43&	47.54\\
    \hline
    CasMVSNet~\cite{gu2020cascade}&56.84	&	76.37&	58.45&	46.26&	55.81&	56.11&	54.06&	58.18&	49.51\\
    \hline
    UCS-Net~\cite{cheng2020deep} & 54.83 & 76.09 & 53.16 & 43.03 & 54.00 & 55.60 & 51.49 & 57.38 & 47.89 \\
    \hline
    $D^2$HC-RMVSNet~\cite{yan2020dense}&	59.20&	74.69&	56.04&	49.42&	60.08&	59.81&{59.61}&	60.04&	53.92 \\
    \hline
    Vis-MVSNet~\cite{zhang2020visibility}&60.03 &	77.40&	60.23&	47.07&	63.44&	{62.21}&	57.28&	{60.54}&	52.07\\
    \hline
    AA-RMVSNet~\cite{wei2021aa} & 61.51 & 77.77 &	59.53&	 51.53 &	64.02&	 64.05 &	59.47&	60.85&	54.90\\
    \hline
    \end{tabular}
    }
\label{tab:tnt}
\end{table*}

\subsection{Topics Remaining Unstudied}
So far, learning-based MVS is still a niche field of computer vision and many well-known and widely used commercial softwares still apply traditional non-learning algorithms. A very typical problem for learning-based MVS algorithms is lack of valid depth values at low-textured regions and that is the reason why SOTA methods of ETH3D are still non-learning ones. Some attempts use different a priori information, such as surface normal~\cite{kusupati2020normal,liu2020depth} to overcome planar areas. But these attempts highly rely on post-process and are still far from end-to-end.

One unstudied topic is what kind of feature extractors are suitable for MVS, as have been mentioned in Sec.~\ref{sec:fe}. MVS relies on a relatively tricky size of receptive fields.

Evaluation metrics are not reasonable. Since MVS serves as a step of image-based 3D reconstruction, whose final purpose is to construct a mesh model and point clouds are intermediate representations. If a good surface reconstruction algorithm could properly estimate faces, then the lack of points during MVS reconstruction is acceptable.

Some researchers have noticed the gap between depth maps and point clouds and want to construct a unified framework for training and evaluation. For example, merging depth fusion into end-to-end training so the loss is directly computed from point clouds. A major problem preventing doing so is how to turn geometric consistency checking differentiable. Some attempts are done, such as Point-MVSNet~\cite{chen2019point}, but the results are not satisfying enough.

Another task is to encode multimodal information into MVS networks, e.g., semantics. In this way, some challenging areas can be dealt individually, such as omitting points with the semantic label of sky and interpolating points within the region labelled as ground. A suitable dataset is also required. 

Last but not least, the number of available datasets and the diversity of data are also quite limited. 

\section{Conclusion}
In this paper, several aspects of learning-based MVS algorithms are covered, including post-processes, plane sweep, relevant datasets and network modules. Besides, comparative results and observations are presented.
Generally speaking, deep-learning-based MVS is still under development and the community is rather tiny compared to other computer vision tasks.




{\small
\bibliographystyle{ieee_fullname}
\bibliography{egbib}
}

\end{document}